\documentclass{article}

\usepackage{microtype}
\usepackage{graphicx}
\usepackage{subfigure}
\usepackage{booktabs} % for professional tables
\usepackage{hyperref}

\usepackage[accepted]{icml2023}

\usepackage{amsmath}
\usepackage{amssymb}
\usepackage{mathtools}
\usepackage{amsthm}
\usepackage{url}
\usepackage{mathrsfs,algpseudocode}

\usepackage{multirow}
\usepackage{graphicx}
\usepackage{wrapfig}

\usepackage{enumitem}
\setlist[itemize]{leftmargin=*}

\usepackage[capitalize,noabbrev]{cleveref}
\usepackage{balance}

\theoremstyle{plain}

\theoremstyle{definition}

\theoremstyle{remark}

\usepackage[textsize=tiny]{todonotes}

\usepackage{setspace}

\icmltitlerunning{Vehicle-Infrastructure Cooperative 3D Object Detection via Feature Flow Prediction}

\begin{document}

\twocolumn[
\icmltitle{Vehicle-Infrastructure Cooperative 3D Object Detection \\ via Feature Flow Prediction}



\icmlsetsymbol{equal}{*}

\begin{icmlauthorlist}
\icmlauthor{Haibao Yu}{lll,ttt}
\icmlauthor{Yingjuan Tang}{bbb,ttt}
\icmlauthor{Enze Xie}{lll}
\icmlauthor{Jilei Mao}{ttt}
\icmlauthor{Jirui Yuan}{ttt}
\icmlauthor{Ping Luo}{lll}
\icmlauthor{Zaiqing Nie}{ttt}
\end{icmlauthorlist}

\icmlaffiliation{lll}{University of Hong Kong}
\icmlaffiliation{ttt}{AIR, Tsinghua University}
\icmlaffiliation{bbb}{Beijing Institute of Technology}

\icmlcorrespondingauthor{Ping Luo}{pluo@cs.hku.hk}
\icmlcorrespondingauthor{Zaiqing Nie}{zaiqing@air.tsinghu.edu.cn}

\icmlkeywords{Machine Learning, ICML}

\vskip 0.3in
]



\printAffiliationsAndNotice{}  % leave blank if no need to mention equal contribution

\begin{abstract}
Cooperatively utilizing both ego-vehicle and infrastructure sensor data can significantly enhance autonomous driving perception abilities. However, temporal asynchrony and limited wireless communication in traffic environments can lead to fusion misalignment and impact detection performance. This paper proposes Feature Flow Net (FFNet), a novel cooperative detection framework that uses a feature flow prediction module to address these issues in vehicle-infrastructure cooperative 3D object detection. Rather than transmitting feature maps extracted from still-images, FFNet transmits feature flow, which leverages the temporal coherence of sequential infrastructure frames to predict future features and compensate for asynchrony. Additionally, we introduce a self-supervised approach to enable FFNet to generate feature flow with feature prediction ability. Experimental results demonstrate that our proposed method outperforms existing cooperative detection methods while requiring no more than 1/10 transmission cost of raw data on the DAIR-V2X dataset when temporal asynchrony exceeds 200$ms$. 
The code is available at \href{https://github.com/haibao-yu/FFNet-VIC3D}{https://github.com/haibao-yu/FFNet-VIC3D}.
\end{abstract}

%%%%%%%%% Introduction
%%%%%%%%%%%%%%%%%%%%%%%%%%%%%%%%%%%%%%%%%%%%%%%%%%%%%%%%%%%%%%%%%%%%
%%%%%%%%%%%%%%%%%%%%%%%%%%%%%%%%%%%%%%%%%%%%%%%%%%%%%%%%%%%%%%%%%%%%
\section{Introduction}
\begin{figure*}
  \centering
  \includegraphics[width=1\linewidth]{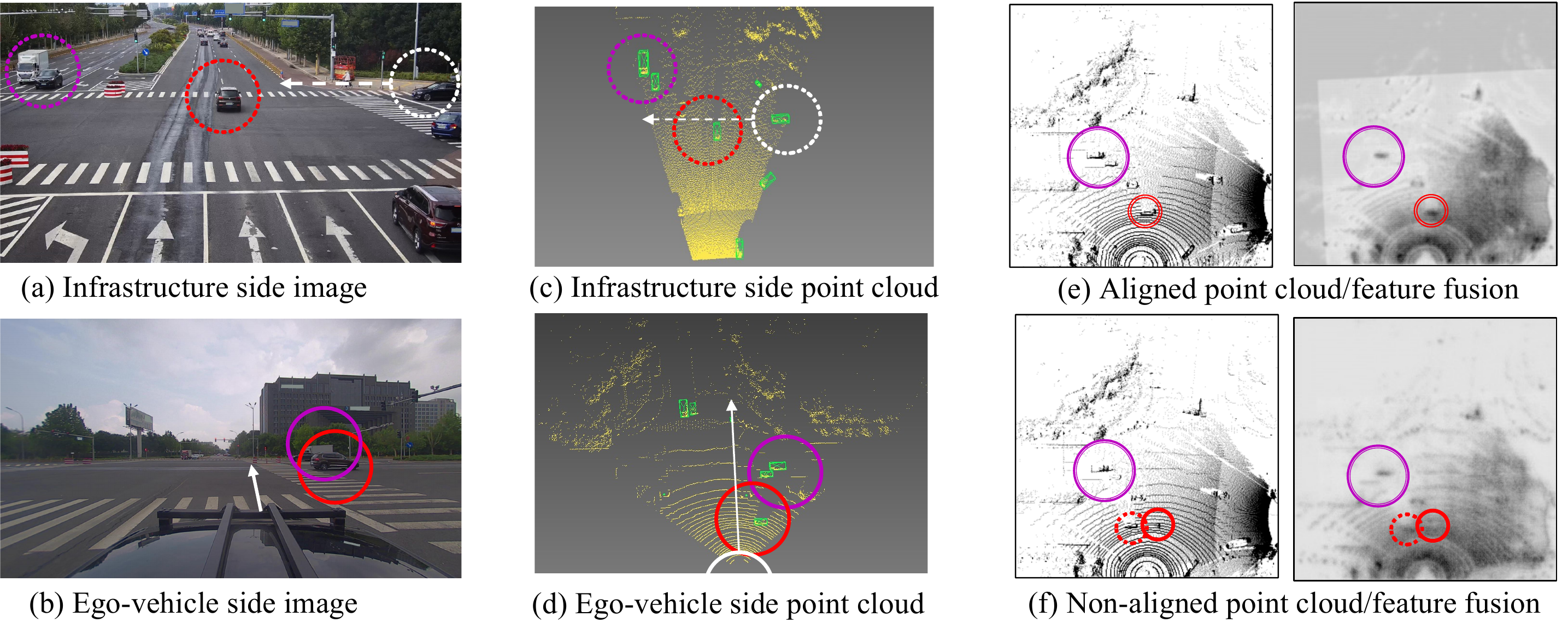}\label{fig: the main idea of FFNet}
  \vspace{-10pt}
  \caption{\textbf{Fusion with Aligned Data and Non-Aligned Data.} (a)-(d) Images and Point clouds Captured from Infrastructure and Vehicle Sensors. The objects in circles with same color share the same object IDs. The object in the white circle indicates the ego vehicle.
  (e) Left: Fusion with Aligned Point Clouds.
  Right: Fusion with Aligned Features. 
    In the absence of temporal asynchrony, infrastructure data and vehicle data depict the same scene simultaneously but with different perspectives. Thus, when we transform infrastructure data (raw data or feature maps) into the same coordinate system as the vehicle data, the data elements in the same position correspond to the encoding of the same objects. Consequently, direct fusion of the two-side data does not result in any fusion misalignment, as depicted in the red circle. 
  (f) Left: Fusion with Non-aligned Point Clouds.
  Right: Fusion with Non-aligned Features. 
  There is a communication latency after broadcasting the infrastructure data. Consequently, dynamic objects (like the object in the red circle) can move, and the data elements in the same position may no longer correspond to the encoding of the same objects, leading to fusion errors when directly fusing the two-side data. This non-alignment of point clouds or features may produce false detection result.
  }
\end{figure*}

3D object detection is a critical task in autonomous driving, providing accurate location and classification information about surrounding obstacles. Traditional 3D object detection utilizes onboard sensor data from the ego vehicle to generate 3D outputs. 
However, with the limited perception field of ego-vehicle sensors, this solution often fails in the blind or long-range zones, affecting downstream prediction and planning tasks and resulting in severe safety problems.
Recently, vehicle-infrastructure cooperative autonomous driving has attracted much attention to address these safety challenges~\cite{yu2022dairv2x}. Infrastructure sensors like cameras and LiDARs are commonly installed much higher than ego vehicles, providing a broader field of view. 
Utilizing extra infrastructure sensor data can effectively obtain more meaningful information and improve autonomous driving perception ability. This paper focuses on solving the vehicle-infrastructure cooperative 3D (VIC3D) object detection problem with point clouds as inputs.

The VIC3D problem can be formulated as a multi-sensor detection problem under constrained communication bandwidth.
This problem presents two main challenges: first, the data captured by ego-vehicle sensors and received from infrastructure devices have asynchronous timestamps; second, communication bandwidth between two-side devices is limited.
Recent studies~\cite{yu2022dairv2x,hu2022where2comm,xu2022v2x} have attempted to address this problem and proposed three major fusion frameworks for cooperative detection: 
early fusion, late fusion, and middle fusion. 
Early fusion involves transmitting raw data like raw point clouds, while late fusion uses detection outputs for object-level fusion. 
Middle fusion utilizes intermediate-level features for feature fusion, striking a balance between preserving valuable information and reducing redundant transmission. 
However, existing middle-fusion solutions~\cite{hu2022where2comm,xu2022v2x} overlook the challenge of temporal asynchrony explicitly, resulting in fusion misalignment that affects the detection results, as depicted in Figure~\ref{fig: the main idea of FFNet}.

In this work, we propose Feature Flow Net (FFNet), a novel cooperative detection framework for vehicle-infrastructure cooperative 3D (VIC3D) object detection. 
FFNet is designed to address the temporal asynchrony between infrastructure and ego-vehicle sensor data that can cause fusion misalignment and impact detection performance. As shown in Figure \ref{fig: FFNet}, FFNet consists of several steps, including generating feature flow from sequential infrastructure frames, transmitting the compressed feature flow, decompressing the feature flow, and reconstructing infrastructure features to align with the timestamp of ego-vehicle sensor data for detection output. The critical module in FFNet is feature flow, which acts as a feature prediction function to generate alignment with ego-vehicle features and eliminate fusion errors caused by temporal asynchrony. Moreover, feature flow is an intermediate-level data that can be compressed, reducing transmission costs.
A self-supervised approach is also proposed to train the feature flow generator by constructing ground truth features to optimize feature flow prediction. 
We extract rich temporal correlations from raw frames embedded in feature flow to predict future infrastructure features and align them with vehicle features, making it highly suitable for addressing the temporal asynchrony challenge in VIC3D. 
To the best of our knowledge, this is the first time utilizing feature flow for multi-sensor object detection to address the challenge of temporal misalignment and using raw frames to enhance feature prediction accuracy.

We conduct an extensive evaluation of the proposed FFNet framework on the DAIR-V2X dataset~\cite{yu2022dairv2x}, which contains real-world driving scenarios. In order to demonstrate the effectiveness of FFNet, we compare its performance with several existing cooperative detection methods, including V2VNet~\cite{wang2020v2vnet} and DiscoNet~\cite{li2021learning}, by implementing them and conducting experiments under the same settings. Our results demonstrate that FFNet can successfully address the challenge of temporal asynchrony by accurately predicting future features at different timestamps, with the ability to overcome various latencies ranging from 100$ms$ to 500$ms$.
Notably, our proposed FFNet outperforms all existing cooperative detection methods and achieves state-of-the-art performance while requiring less than 1/10 of the transmission cost of raw data, particularly when the latency exceeds 200$ms$.

The main contributions are organized as followings.
\begin{itemize}[itemsep=2pt,topsep=0pt,parsep=0pt]
    \item We propose a novel intermediate-level framework for VIC3D problem, called Feature Flow Net (FFNet), which utilizes feature flow for cooperative detection. The FFNet generates future features to address the temporal asynchrony challenge and reduces transmission costs while retaining valuable information for cooperative detection.
    \item We introduce a self-supervised approach to train the feature flow generator in FFNet, which successfully predicts future features to align with ego-vehicle sensor data and mitigate temporal fusion errors in various latencies.
    \item We evaluate the proposed FFNet framework on the real-world DAIR-V2X dataset, demonstrating superior performance compared to other state-of-the-art intermediate-level methods. Furthermore, FFNet surpasses all cooperative detection methods when latency exceeds 200$ms$.
\end{itemize}

%%%%%%%% Related Work
%%%%%%%%%%%%%%%%%%%%%%%%%%%%%%%%%%%%%%%%%%%%%%%%%%%%%%%%%%%%%%%%%%%
%%%%%%%%%%%%%%%%%%%%%%%%%%%%%%%%%%%%%%%%%%%%%%%%%%%%%%%%%%%%%%%%%%%
\section{Related Work}\label{sec: related work}
\paragraph{Egocentric 3D Object Detection.} 
Perceiving objects, especially 3D obstacles in the road environment, is a fundamental task in egocentric autonomous driving. 
Egocentric 3D object detection can be classified into three categories based on sensor types: camera-based methods, LiDAR-based methods, and multi-sensor-based methods. 
Camera-based methods, such as FCOS3D~\cite{wang2021fcos3d}, directly detect 3D bounding boxes from a single image. Other camera-based methods, such as BEVformer~\cite{li2022bevformer} and M$^2$BEV~\cite{xie2022m}, project 2D images onto a bird's-eye view (BEV) to conduct multi-camera joint 3D detection. LiDAR-based methods, such as VoxelNet~\cite{zhou2018voxelnet}, SECOND~\cite{yang2019std}, and PointPillars~\cite{lang2019pointpillars}, divide the LiDAR point cloud into voxels or pillars and extract features from them. Multi-sensor-based methods, such as PointPainting~\cite{vora2020pointpainting} and BEV-Fusion~\cite{liu2022bevfusion}, utilize both camera and LiDAR data.
In contrast to these methods for single-vehicle view object detection, our proposed method focuses on cooperative detection. It utilizes both infrastructure and vehicle sensor data to overcome the perception limitations of single-vehicle view detection.

\paragraph{VIC3D Object Detection.}
With the development of V2X communication~\cite{hobert2015enhancements}, utilizing information from the road environment has attracted much attention. 
Some works use information from other vehicles to broaden the perception field. 
V2VNet~\cite{wang2020v2vnet} is a pioneering work in multi-vehicle 3D perception and provides a feature fusion framework to achieve performance-transmission trade-off. DiscoNet~\cite{li2021learning} applies distillation in the feature fusion training. 
V2X-ViT~\cite{xu2022v2x} introduces the vision transformer to fuse information across on-road agents. 
V2X-Sim~\cite{li2022v2x} and OPV2V~\cite{xu2021opv2v} are two simulated datasets for multi-vehicle cooperative perception research. 
Some works~\cite{valiente2019controlling, cui2022coopernaut} integrate infrastructure data for end-to-end autonomous driving. 
Some works use infrastructure data to improve 3D perception ability.
DAIR-V2X~\cite{yu2022dairv2x} is a pioneering work in vehicle-infrastructure cooperative 3D object detection. 
It releases a large-scale real-world V2X dataset and introduces the VIC3D object detection task. 
However, it only provides simple early and late fusion baselines. 
\cite{hu2022where2comm, arnold2020cooperative} focus on transmitting feature maps for cooperative detection, but these works do not consider the temporal asynchrony and fusion errors.
In contrast, our approach overcomes the challenge of temporal asynchrony and reduces transmission costs. 
We propose a feature flow prediction method that is different from \cite{lei2022latency}, which focuses on integrating per-frame features from other vehicles. 
Our approach addresses the challenge of temporal asynchrony by predicting future feature and compensating for the latency, resulting in improved detection performance.

\paragraph{Feature Flow.}
Flow is a concept originating from mathematics, which formalizes the idea of the motion of points over time~\cite{deville2012mathematical}. 
It has been successfully applied to many computer vision tasks, such as optical flow~\cite{beauchemin1995computation}, scene flow~\cite{menze2015object}, and video recognition~\cite{zhu2017deep}. 
As a concept extended from optical flow~\cite{horn1981determining}, feature flow describes the changing of feature maps over time, and it has been widely used in various video understanding tasks.
Zhu et al.\cite{zhu2017flow} propose a flow-guided feature aggregation to improve video detection accuracy. 
In this paper, we introduce the feature flow for feature prediction to overcome the challenge of temporal asynchrony in VIC3D object detection.

%%%%%%%%% Feature Flow Net
%%%%%%%%%%%%%%%%%%%%%%%%%%%%%%%%%%%%%%%%%%%%%%%%%%%%%%%%%%%%%%%%%%%%
%%%%%%%%%%%%%%%%%%%%%%%%%%%%%%%%%%%%%%%%%%%%%%%%%%%%%%%%%%%%%%%%%%%%
\section{Method}
\begin{figure*}[ht]
  \centering
  \includegraphics[width=1\linewidth]{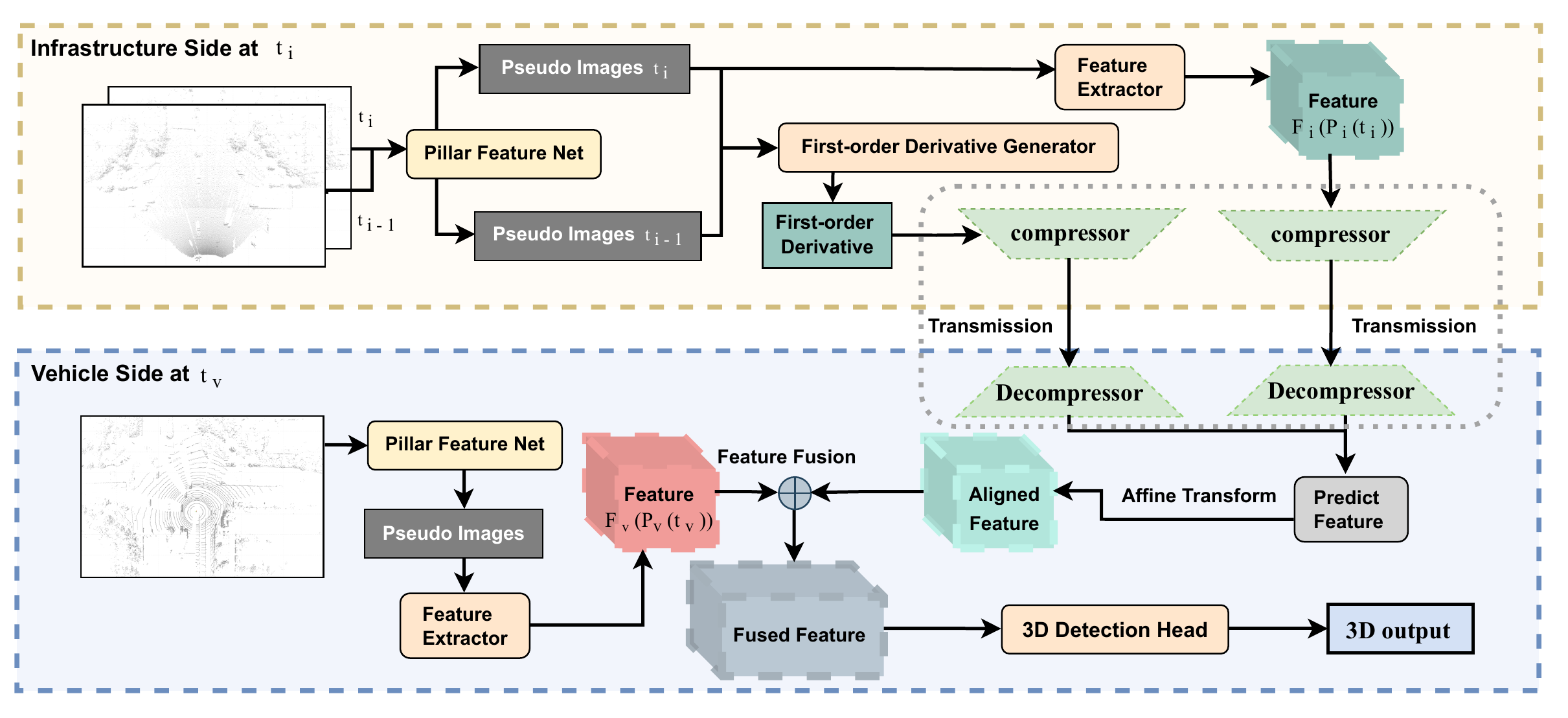}
  \vspace{-10pt}
  \caption{FFNet Overview.
  In the infrastructure system, the feature extractor and the first-order derivative generator make up the feature flow generator. 
  The feature flow is linearly represented with a feature and a first-order derivative as shown in Equation~\ref{eq: first-order of feature flow}, and can be used to predict the future feature.
  The compressors and decompressors focus on reducing the transmission cost while reserving valuable information for cooperative detection. Detailed configurations for FFNet can be found in the Appendix.
  }\label{fig: FFNet}
\end{figure*}

This paper aims to improve vehicle-infrastructure cooperative 3D (VIC3D) object detection while considering practical communication conditions to reduce transmission costs and overcome the temporal asynchrony challenge.
To achieve this goal, we first apply V2VNet~\cite{wang2020v2vnet} for cooperative detection, which is originally used to transmit intermediate features for multi-vehicle perception. We provide details on this basic solution in the Appendix. 
However, this approach ignores temporal asynchrony and may introduce serious fusion misalignment.
To address these issues, we propose a novel intermediate-level framework called Feature Flow Net (FFNet), which includes a feature flow prediction function that generates aligned infrastructure features with vehicle features to compensate for the temporal asynchrony.
Moreover, we introduce a self-supervised approach to train the feature flow generator.
In the following sections, we describe the VIC3D problem in Section~\ref{sec: vic3d detection task}, detail the inference of our solution in Section~\ref{sec: FFNet}, and explain the training in Section~\ref{sec: semi-supervised learning}.
We also compare different possible solutions in the Appendix.

\subsection{VIC3D Object Detection}\label{sec: vic3d detection task}
The VIC3D object detection aims to improve the performance of localizing and recognizing the surrounding objects by utilizing both the infrastructure and vehicle sensor data under limited communication conditions. 
This paper focuses on point clouds captured from LiDAR as inputs. 
The input of VIC3D consists of two parts:
\begin{itemize}[itemsep=2pt,topsep=0pt,parsep=0pt]
    \item Point cloud $P_{v}(t_{v})$ captured by the vehicle sensor with timestamp $t_{v}$ as well as its relative pose $M_{v}(t_{v})$, where $P_{v}(\cdot)$ denotes the capturing function of vehicle LiDAR.
    \item Point cloud $P_{i}(t_{i})$ captured by the infrastructure sensor with timestamp $t_{i}$ as well as its relative pose $M_{i}(t_{i})$, where $P_{i}(\cdot)$ denotes the capturing function of infrastructure LiDAR. Previous frames captured by the infrastructure sensor can also be utilized in cooperative detection.
\end{itemize}
Note that the timestamp $t_i$ should be earlier than timestamp $t_v$ since receiving the data through long-range communication from infrastructure devices to vehicle devices requires a significant amount of transmission time. 
Moreover, the latency $(t_v-t_i)$ should be uncertain before receiving the data, as the transmitted data could be obtained by various autonomous driving vehicles in different locations after infrastructure broadcasting.
More illustration of the uncertain latency is provided in the Appendix.

Compared with 3D object detection in single-vehicle autonomous driving, the VIC3D faces more challenges in temporal asynchrony and transmission costs. 
Directly fusing the infrastructure data can cause serious fusion errors and affect the detection performance due to the scene changing and dynamic objects moving. 
Asynchronous phenomena are also demonstrated in the experiment results in Section~\ref{sec: ablation study}. 
Additionally, reducing the amount of transmitted data could effectively shorten the overall latency because the transmission time is positively correlated with the amount of the transmitted data~\cite{fano1961transmission}. 
Therefore, in addition to the detection performance, the VIC3D has another goal: to reduce the transmission cost. 
Here, mean Average Precision (mAP) is used to measure the detection performance, while the Average Byte ($\mathcal{AB}$) is used to measure the transmission cost. 
We detail the two metrics in the Appendix.

\subsection{Feature Flow Net}\label{sec: FFNet}
Our Feature Flow Net (FFNet) provides a new intermediate-level solution for VIC3D. 
This approach transmits the compressed feature flow for cooperative detection, which reduces redundant information in raw data and eliminates the fusion misalignment caused by temporal asynchrony. 
The inference process of FFNet consists of following stages: (1) generating the feature flow, (2) compressing, transmitting, and decompressing the feature flow, and (3) predicting the aligned feature and fusing it with the vehicle feature to generate the detection results. The inference process is illustrated in Figure~\ref{fig: FFNet}.

\paragraph{Feature Flow Concept for VIC3D.}
The concept of flow is based on mathematical models that formalize the idea of the motion of points over time~\cite{deville2012mathematical}. 
Feature flow is a variant that characterizes the intermediate feature changes over time. 
In this paper, we adopt the feature flow as a prediction function to describe the infrastructure feature changes over time in the future. 
Given the current point cloud frame $P_i(t_i)$ and the infrastructure feature extractor $F_i(\cdot)$, the feature flow over the future time $t$ after $t_i$ is defined as:
\begin{equation}
\setlength\abovedisplayskip{0.05cm}
\setlength\belowdisplayskip{0.15cm}
\widetilde{F}_i(t) =
F_i(P_i(t)), t \geq t_i
\end{equation}

After an uncertain transmitting time, some autonomous driving vehicles receive the feature flow. 
Compared with the previous approach of transmitting feature $F_i(P_i(t_i))$ produced from per frames~\cite{wang2020v2vnet}, which lacks temporal and predictive information, feature flow enables the direct prediction of the aligned feature at the timestamp $t_v$ of the vehicle sensor data. 
We fuse this predicted infrastructure feature $\widetilde{F}_i(t_v)$ with the vehicle feature to compensate for the asynchronous time and eliminate the fusion error caused by temporal asynchrony. 
Note that we generate the feature flow to obtain temporal prediction ability on the infrastructure side rather than the vehicle side. 
This design is because we can extract richer temporal correlation from raw sequential frames than from information-reduction features. 
We also implement the comparison of the extraction on different sides in the Appendix. 
We will introduce how to implement the feature flow in the next part.

\paragraph{Feature Flow Generation.}
Two issues need to be addressed in order to apply the feature flow to transmission and cooperative detection: expressing and transmitting the continuous feature flow changes over time, and enabling the feature flow with prediction ability. 
Considering that the time interval $t_v\rightarrow t_i$ is generally short, we address the expressing issue by using the simplest first-order expansion to represent the continuous feature flow over time, which takes the form of Equation~\eqref{eq: first-order of feature flow}, 
\begin{equation}\label{eq: first-order of feature flow}
    \setlength\abovedisplayskip{0.05cm}
    \setlength\belowdisplayskip{0.15cm}
    \widetilde{F}_i(t_i+\Delta t) \approx F_{i}(P_i(t_i)) + \Delta t * \widetilde{F}_i^{'}(t_i),
\end{equation}
where $\widetilde{F}i^{'}(t_i)$ denotes the first-order derivative of the feature flow and $\Delta t$ denotes a short time period in the future. 
Thus, we only need to obtain the feature $F{i}(P_i(t_i))$ and the first-order derivative of the feature flow $\widetilde{F}_i^{'}(t_i)$ to obtain an approximate representation of the feature flow.
When an autonomous driving vehicle receives the feature $F_{i}(P_i(t_i))$ and the first-order derivative $\widetilde{F}_i^{'}(t_i)$ after an uncertain latency, we can generate the infrastructure feature aligned with the vehicle sensor data, and the prediction operation consumes minor computation because it only needs linear calculation. 
To enable the feature flow with prediction ability, we use a network to extract the first-order derivative of the feature flow $\widetilde{F}_i^{'}(t_i)$ from the historical infrastructure frames ${I_i(t_i-N+1), \cdots, I_i(t_i-1), I_i(t_i)}$. 
Generally, the larger $N$ will generate more accurate estimations.
In this paper, we take $N$ as two and use two consecutive infrastructure frames $P_i(t_i-1)$ and $P_i(t_i)$.

Specifically, we first use the Pillar Feature Net~\cite{lang2019pointpillars} to convert the two consecutive point clouds into two pseudo-images with a bird-eye view (BEV) and with the size of $[384, 288, 288]$. Then, we concatenate the two BEV pseudo-images into the size of $[768, 288, 288]$, and input the concatenated pseudo-images into a 13-layer Backbone and a 3-layer FPN (Feature Pyramid Network), as in SECOND~\cite{yan2018second}, to generate the estimated first-order derivative $\widetilde{F}_i^{'}(t_i)$ with the size of $[364, 288, 288]$. Finally, with the feature $F_{i}(P_i(t_i))$ and the estimated first-order derivative $\widetilde{F}_i^{'}(t_i)$, we can represent the feature flow with prediction ability. The detailed network configuration is provided in the Appendix.

\paragraph{Compression, Transmission and Decompression.}
The transmission cost of the feature flow in its original form is substantial, amounting to up to 243Mb~\footnote{This transmission involves 384$\times$288$\times$288$\times$2 floating point numbers, which equates to 243Mb.}. 
In order to eliminate redundant information and reduce the transmission cost further, we apply two compressors to the feature $F_{i}(P_i(t_i))$ and the derivative $\widetilde{F}_i^{'}(t_i)$, compressing them from size $[384, 288, 288]$ to $[12, 36, 36]$ using three Conv-Bn-ReLU blocks in each compressor.
The size of the timestamp and calibration file is small and cannot be ignored. 
Therefore, the compressed feature flow only requires 0.12Mb per transmission, which is $1/32$$\times$$1/8$$\times$$1/8$ of the original feature flow. 
We broadcast the compressed feature flow along with the corresponding timestamp and calibration file on the infrastructure side.
Upon receiving the compressed feature flow, the vehicle uses two decompressors, each composed of three Deconv-Bn-ReLU blocks, to decompress the compressed feature and compressed first-order derivatives from size $[384/32, 288/8, 288/8]$ to the original size $[384, 288, 288]$.

\paragraph{Vehicle-Infrastructure Feature Fusion.}
We employ a Pillar Feature Net~\cite{lang2019pointpillars} and a feature extractor to generate the vehicle feature from the vehicle point cloud $P_v(t_v)$, which is captured at timestamp $t_v$. The decompressed feature flow is then utilized to predict the infrastructure feature at timestamp $t_v$, aligned with the vehicle feature, as follows:
\begin{equation}
\setlength\abovedisplayskip{0.05cm}
\setlength\belowdisplayskip{0.15cm}
\widetilde{F}i(t_v) \approx F{i}(P_i(t_i)) + (t_v-t_i) * \widetilde{F}_i^{'}(t_i).
\end{equation}
This linear prediction operation effectively compensates for uncertain latency and requires minimal computation. 
The predicted feature $\widetilde{F}_i(t_v)$ is then transformed into the vehicle coordinate system using the corresponding calibration files. 
The bird's-eye view of the infrastructure and vehicle features are obtained, both at the vehicle coordinate system, while preserving spatial alignment. 
The feature located outside the vehicle's interest area is discarded for the infrastructure feature, and empty locations are padded with zero elements.

Subsequently, we concatenate the infrastructure feature with the vehicle feature and employ a Conv-Bn-Relu block to fuse the concatenated features to obtain the vehicle-infrastructure fused feature. 
Finally, we input the fused feature into a 3D detection head, utilizing the Single Shot Detector (SSD)~\cite{liu2016ssd} setup as the 3D object detection head, to generate 3D outputs for more accurate localization and recognition. 
The experimental results indicate that the infrastructure feature flow significantly enhances the detection ability.

\subsection{Training Feature Flow Net}\label{sec: semi-supervised learning}
The FFNet training consists of two stages: training a basic fusion framework in an end-to-end way and then using a self-supervised learning to train the feature flow generator.

\paragraph{Training Basic Fusion Framework.}
In the first stage, we train a basic fusion framework in an end-to-end manner without considering latency. 
The objective of this stage is to train FFNet to fuse the infrastructure feature with the vehicle feature to enhance detection performance. 
Specifically, we train FFNet using cooperative data and annotations obtained from both the vehicle and the infrastructure. 
The localization regression and object classification loss functions used in SECOND~\cite{yan2018second} are applied in this stage. 
To speed up training, we mask the first-order derivative generation and use only one infrastructure point cloud $P_i(t_i)$. 
At the end of this stage, all the modules in FFNet are trained except for the first-order derivative generator.

\paragraph{Training Feature Flow Generator.}
\begin{figure}
  \centering
  \includegraphics[width=0.45\textwidth]{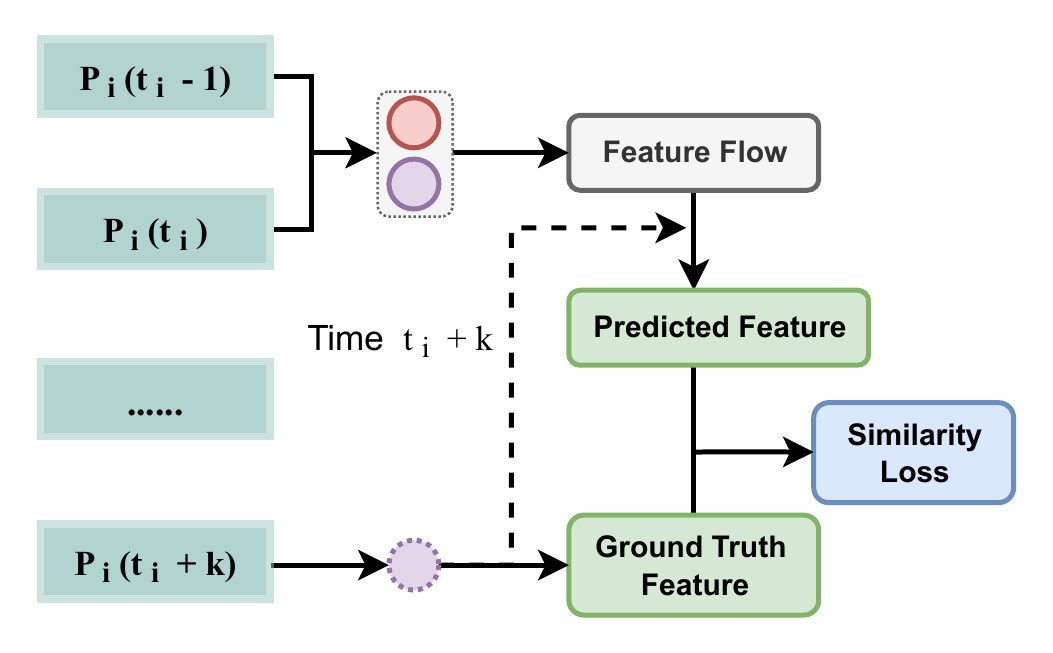}
  \vspace{-5pt}
  \caption{Illustration of training a feature flow generator using self-supervised learning and similarity loss. The upper red circle represents the first-order derivative generator, while the lower purple circles with solid and dashed lines share the same infrastructure feature extractor.
  Feature flow predicts the feature at $t_i+k$.
  }\label{fig: ffgenerator-training}
\end{figure}
In the second stage, we use self-supervised learning to train the feature flow generator by exploiting the temporal correlations in infrastructure sequences, as shown in Figure~\ref{fig: ffgenerator-training}.
The idea is to generate the feature flow, which contains information about the movement of objects over time, and then use it to predict the feature at a future time. 
We construct the ground truth feature by using nearby infrastructure frames that do not require any manual annotations. 
Specifically, we generate training frame pairs $\mathcal{D}={d_{t_i,k}=(P_i(t_i-1), P_i(t_i), P_i(t_i+k))}$, where $P_i(t_i-1)$ and $P_i(t_i)$ are two consecutive infrastructure point cloud frames, and $P_i(t_i+k)$ is the $(k+1)$-th frame after $P_i(t_i)$. 
For each pair $d_{t_i,k}$ in $\mathcal{D}$:
\vspace{-0.1cm}
\begin{itemize}[itemsep=2pt,topsep=0pt,parsep=0pt]
    \item we input $P_i(t_i-1)$ and $P_i(t_i)$ into the feature flow generator to generate the feature flow, which is composed of $F_{i}(P_i(t_i))$ and the estimated first-order derivative $\widetilde{F}_i^{'}(t_i)$. 
    \item We then use the feature flow to predict the feature at time $t_i+k$ as
    \begin{equation}
        \setlength\abovedisplayskip{0.05cm}
        \setlength\belowdisplayskip{0.15cm}
        \widetilde{F}_{i}(t_i+k) \approx F_{i}(P_i(t_i)) + |(t_i+k)-t_i| * \widetilde{F}_i^{'}(t_i).
    \end{equation}
    \item We use the infrastructure feature extractor $F_i(\cdot)$ to extract the feature $F_i(P_i(t_i+k))$ from $P_i(t_i+k)$ as the ground truth feature to supervise the feature flow prediction.
\end{itemize}

We construct the loss function to optimize the feature flow generator. 
The objective is to generate the feature flow and use it to predict $\widetilde{F}_i(t_i+k)$ as close as possible to $F_i(P_i(t_i+k))$. 
We use the cosine similarity to measure the similarity between the predicted feature and the ground truth feature as
\begin{equation}
    \setlength\abovedisplayskip{0.05cm}
    \setlength\belowdisplayskip{0.15cm}
    similarity = \frac{\widetilde{F}_i(t_i+k) \odot F_i(P_i(t_i+k))}{||\widetilde{F}_i(t_i+k)||_2*||F_i(P_i(t_i+k))||_2},
\end{equation}
where $\odot$ denotes the inner product, $*$ denotes the scalar multiplication, and $||\cdot||_2$ denotes the L2 norm. 
We use this similarity as the loss function to train the feature flow generator as
\begin{equation}\label{Eq: loss}
    \setlength\abovedisplayskip{0.05cm}
    \setlength\belowdisplayskip{0.15cm}
    \mathcal{L}(\mathcal{D},\theta) = \sum_{d_{t_i,k}\in \mathcal{D}} (1-\frac{\widetilde{F}_i(t_i+k) \odot F_i(P_i(t_i+k))}{||\widetilde{F}_i(t_i+k)||_2*||F_i(P_i(t_i+k))||_2}),
\end{equation}
where $\theta$ is the parameter of the feature flow generator, and we only update the parameters in first-order derivative generator $\widetilde{F}_i^{'}(\cdot)$ and frozen other parameters.
To train the feature flow generator, we use the Adam optimizer~\cite{kingma2014adam} to minimize the loss function with the batch size of 2, the learning rate of 0.001, and the total epoch of 10. 

\paragraph{Remark.} 
It is important to note that cosine similarity is unaffected by the magnitudes of the input tensors. For instance, the input tensors $[1, 2, 3]$ and $[2, 4, 6]$ achieve a maximal cosine similarity value of $1$, even though they have different values. 
Therefore, to adjust the magnitudes of $\widetilde{F}_i(t_i+k)$, we require a scale transformation. We propose using $||F_i(P_i(t_i))||_1 / ||\widetilde{F}_i(t_i+k)||_1$ as the scaling factor, where $||\cdot||_1$ denotes the L1 norm.

%%%%%%%%% Experiments
%%%%%%%%%%%%%%%%%%%%%%%%%%%%%%%%%%%%%%%%%%%%%%%%%%%%%%%%%%%%%%%%%%%%
%%%%%%%%%%%%%%%%%%%%%%%%%%%%%%%%%%%%%%%%%%%%%%%%%%%%%%%%%%%%%%%%%%%%
\section{Experiments}\label{sec: experiments}
\begin{table*}[ht]
\small
\centering
\caption{\textbf{VIC3D Object Detection Results with FFNet and Different Fusion Methods.} $\mathcal{AB}$ denotes the average byte used to measure the transmission cost. "/" denotes no latency for non-fusion methods. "-" denotes no information provided. FFNet outperforms non-fusion methods by up to 10.90\% mAP@BEV (IoU=0.5) and 10.96\% mAP@BEV (IoU=0.5) in 100$ms$ and 200$ms$ latency, respectively. And FFNet outperforms all other middle-fusion methods including DiscoNet and V2VNet while requiring the same transmission cost. FFNet also outperforms all other fusion methods when the latency reaches 200$ms$. Notably, FFNet achieves more than 2\% mAP@BEV (IoU=0.5) improvement over early fusion in 200$ms$ latency while requiring no more than $1/10$ transmission cost.
}\label{tab: experiment performance comparison}
\scalebox{1.0}{
\begin{tabular}{ccc|lcllc}
\hline
\hline
\multirow{2}{*}{Model} & \multirow{2}{*}{FusionType} & \multirow{2}{*}{Latency ($ms$)} & \multicolumn{2}{c}{mAP@3D~$\uparrow$}                & \multicolumn{2}{c}{mAP@BEV~$\uparrow$} & \multirow{2}{*}{$\mathcal{AB}$~(Byte)~$\downarrow$} \\ \cline{4-7} 
&   &  & IoU=0.5 & \multicolumn{1}{l}{IoU=0.7} & IoU=0.5    & IoU=0.7    &    \\
\hline \hline
PointPillars~\cite{lang2019pointpillars}  & non-fusion & /              & 48.06   & -                   & 52.24      & -      & 0    \\
\hline \hline
Early Fusion  & early & 100              & 57.35   & -                   & 64.06      & -      & 1.4$\times 10^6$    \\
TCLF~\cite{yu2022dairv2x}     & late     & 100              & 40.79   & -                   & 46.80      & -      & 5.4$\times 10^2$     \\
DiscoNet~\cite{li2021learning}  & middle    & 100              & 52.83   & 29.19                  & 61.25      & 50.11      & 1.2$\times 10^5$   \\
V2VNet~\cite{wang2020v2vnet}    & middle    & 100              & 52.02   & 28.54                   &  60.78     & 50.02      & 1.2$\times 10^5$   \\
\textbf{FFNet (Ours)}   & middle & 100   & 55.48   & 31.50                    & \textbf{63.14} (\textcolor{red}{+10.90})     & 54.28      & 1.2$\times 10^5$   
\\ \hline \hline                       
Early Fusion & early & 200              & 54.63   & -                   & 61.08      & -      & 1.4$\times 10^6$     \\
TCLF~\cite{yu2022dairv2x}     & late     & 200              & 36.72   & -                   & 41.67      & -      & 5.1$\times 10^2$     \\
DiscoNet~\cite{li2021learning}  & middle    & 200              & 50.76   & 28.57                  & 58.20      & 48.90      & 1.2$\times 10^5$   \\
V2VNet~\cite{wang2020v2vnet}    & middle    & 200              & 49.67   & 26.96                   & 56.02      & 46.32      & 1.2$\times 10^5$   \\
\textbf{FFNet (Ours)}   & middle & 200 & 55.37   & 31.66                   & \textbf{63.20} (\textcolor{red}{+10.96})      & 54.69      & 1.2$\times 10^5$       \\ \hline \hline
\end{tabular}
}
\label{table_xiaorong}
\end{table*}

In this section, we present our approach to solve VIC3D on the DAIR-V2X dataset~\cite{yu2022dairv2x} using the FFNet and various fusion methods. We compare the experimental results of different fusion methods on various latency. Our proposed FFNet outperforms all other methods, including early fusion, when the latency reaches 200$ms$, while only requiring a transmission cost of no more than 1/10 of that of early fusion.
Furthermore, we investigate how the FFNet overcomes the challenge of temporal asynchrony with our proposed feature flow. Our experimental results show that temporal asynchrony can significantly reduce the performance of the cooperative detection model, but the feature flow can effectively compensate for this drop.
Additionally, we evaluate the robustness of the FFNet to uncertain latency by testing it on different latency. The experimental results show that our FFNet can robustly solve the problem of uncertain latency.
In the Appendix, we compare the performance of the feature flow extraction on different sides, i.e., the infrastructure side and the ego vehicle side. We develop our models using MMDetection3D~\cite{mmdet3d2020}.

\subsection{DAIR-V2X Dataset}
DAIR-V2X~\cite{yu2022dairv2x} is a large-scale vehicle-infrastructure cooperative perception dataset consisting of real-world images and point clouds.
With over 100 scenes and 18,000 data pairs, this dataset captures the infrastructure and vehicle sensors simultaneously at an equipped intersection when an autonomous driving vehicle passes through. 
All frames in this dataset are annotated with 10 3D object classes. 
Furthermore, cooperative 3D annotations with vehicle-infrastructure cooperative view are provided for 9,311 pairs, where each object is labeled with its corresponding category (\textit{Car}, \textit{Bus}, \textit{Truck}, or \textit{Van}). 
The dataset is divided into \textit{train/val/test} sets in a 5:2:3 ratio, with all models evaluated on the \textit{val} set.

\subsection{Comparison with Different Fusion Methods}\label{sec: exp-FFNet}
We compare our proposed FFNet with four different fusion methods: non-fusion (e.g., PointPillars~\cite{lang2019pointpillars}), early fusion, late fusion (e.g., TCLF~\cite{yu2022dairv2x}), middle fusion (e.g., DiscoNet~\cite{li2021learning}) and V2VNet~\cite{li2022v2x}. 
The performance of FFNet and these fusion methods are evaluated under 100$ms$ and 200$ms$ latency, respectively, on the DAIR-V2X dataset~\cite{yu2022dairv2x}. 
The detection performance is measured using KITTI~\cite{geiger2012we} evaluation detection metrics: bird-eye view (BEV) mAP and 3D mAP with 0.5 IoU and 0.7 IoU, respectively. Only objects located in the rectangular area [0, -39.12, 100, 39.12] are considered in the metrics, and only the \textit{Car} class is taken into account. 
Implementation details of FFNet and the fusion methods are provided in the Appendix. Evaluation results are presented in Tab.~\ref{tab: experiment performance comparison}.

\paragraph{Result Analysis.}
Firstly, our proposed FFNet outperforms the non-fusion method PointPillars by 10.90\% BEV-mAP (IoU=0.5) and 10.96\% BEV-mAP (IoU=0.5) in 100$ms$ and 200$ms$ latency, respectively. This result indicates that utilizing infrastructure data can improve 3D detection performance.
Secondly, although late fusion requires little transmission cost, the BEV-mAP (IoU=0.5) of TCLF is much lower than that of FFNet, up to 21.53\% in 200$ms$ latency.
Thirdly, compared with early fusion methods, FFNet achieves similar detection performance in 100$ms$ latency and outperforms more than 2\% mAP in 200$ms$ latency, while it only requires no more than 1/10 transmission cost.
Fourthly, our FFNet achieves the best detection performance with the exact transmission cost as the middle fusion methods. For example, FFNet surpasses DiscoNet by 1.89\% BEV-mAP (IoU=0.5) and 5.0\% BEV-mAP (IoU=0.5) in 100$ms$ and 200$ms$ latency, respectively.
In summary, our proposed FFNet achieves new SOTA on DAIR-V2X when the latency reaches 200$ms$, while only requiring no more than 1/10 transmission cost of raw data. Moreover, it not only surpasses all other intermediate-level methods but also surpasses the early fusion.

\begin{table*}
\small
\centering
\caption{\textbf{Comparison between with and without Feature Prediction.} Compared with no prediction models, FFNet with feature prediction has a significantly lower performance drop when there is communication latency.}\label{tab: ablation study of feature prediction}
\scalebox{1.0}{
\begin{tabular}{cc|lcllc}
\hline
\hline
\multirow{2}{*}{Model}  & \multirow{2}{*}{Latency (ms)} & \multicolumn{2}{c}{mAP@3D~$\uparrow$}                & \multicolumn{2}{c}{mAP@BEV~$\uparrow$} & \multirow{2}{*}{AB~(Byte)~$\downarrow$} \\ \cline{3-6} 
 &  & IoU=0.5 & \multicolumn{1}{l}{IoU=0.7} & IoU=0.5    & IoU=0.7    &    \\
\hline \hline
FFNet   & 0              & 55.81   & 30.23                  & 63.54      & 54.16      & 1.2$\times 10^5$    \\
FFNet (without prediction)         & 0              & 55.81   & 30.23                  & 63.54      & 54.16     & 6.2$\times 10^4$     \\
FFNet-V2 (without prediction)      & 0              & 55.78   & 30.22                  & 64.23      & 55.00      & 1.2$\times 10^5$   \\
\hline \hline
FFNet   & 200              & 55.37   & 31.66                   & 63.20 (\textcolor{red}{-0.34})       & 54.69      & 1.2$\times 10^5$    \\
FFNet (without prediction)         & 200              & 50.27   & 27.57                  & 57.93 (\textcolor{red}{-5.61})      & 48.16     & 6.2$\times 10^4$     \\
FFNet-V2 (without prediction)      & 200              & 49.90   & 27.33                  & 58.00 (\textcolor{red}{-6.23})      & 48.22      & 1.2$\times 10^5$   \\
\hline \hline
\end{tabular}
}
\label{table_xiaorong}
\end{table*}

\subsection{Ablation Study}\label{sec: ablation study}
We conducted a series of experiments to demonstrate the effectiveness of the feature flow module in overcoming the temporal asynchrony challenge, as well as to show that FFNet performs robustly under various latencies.

\paragraph{Feature prediction can well solve temporal asynchrony.}
We first evaluated the performance of FFNet under 0$ms$ latency and 200$ms$ latency on DAIR-V2X, where 0$ms$ latency indicates no temporal asynchrony between infrastructure data and vehicle data. 
We then removed the prediction module from FFNet to directly fuse the infrastructure feature without temporal compensation, which we refer to as FFNet (without prediction). We evaluated FFNet (without prediction) under both 0$ms$ and 200$ms$ latency. 
Since FFNet (without prediction) does not require the transmission of the first-order derivative of the feature flow, it only requires half the transmission cost of FFNet. To make a fair comparison, we also trained another version of FFNet, called FFNet-V2, which compressed the feature flow from (384, 288, 288) to (384/16, 288/8, 288/8). Thus, FFNet-V2 (without prediction) has the same transmission cost as FFNet. We evaluated FFNet-V2 (without prediction) under both 0$ms$ and 200$ms$ latency as well. The evaluation results are presented in Table~\ref{tab: ablation study of feature prediction}.

As shown in Table~\ref{tab: ablation study of feature prediction}, FFNet (without prediction) and FFNet-V2 (without prediction) both exhibit a significant performance drop when there is a 200$ms$ latency. For example, FFNet (without prediction) experiences a 5.61\% BEV-mAP (IoU=0.5) drop in 200$ms$ latency compared to 0$ms$ latency. Although FFNet-V2 (without prediction) performs slightly better than FFNet and FFNet (without prediction) in 0$ms$ latency, FFNet significantly surpasses FFNet-V2 (without prediction) in 200$ms$ latency. These results demonstrate that temporal asynchrony can significantly impact performance if we directly fuse the infrastructure feature, and that our feature prediction module can effectively compensate for the performance drop caused by temporal asynchrony.
 
\paragraph{FFNet is robust to uncertain latency.}
\begin{figure}
\centering
\vspace{-0.1cm}
\includegraphics[width=0.40\textwidth]{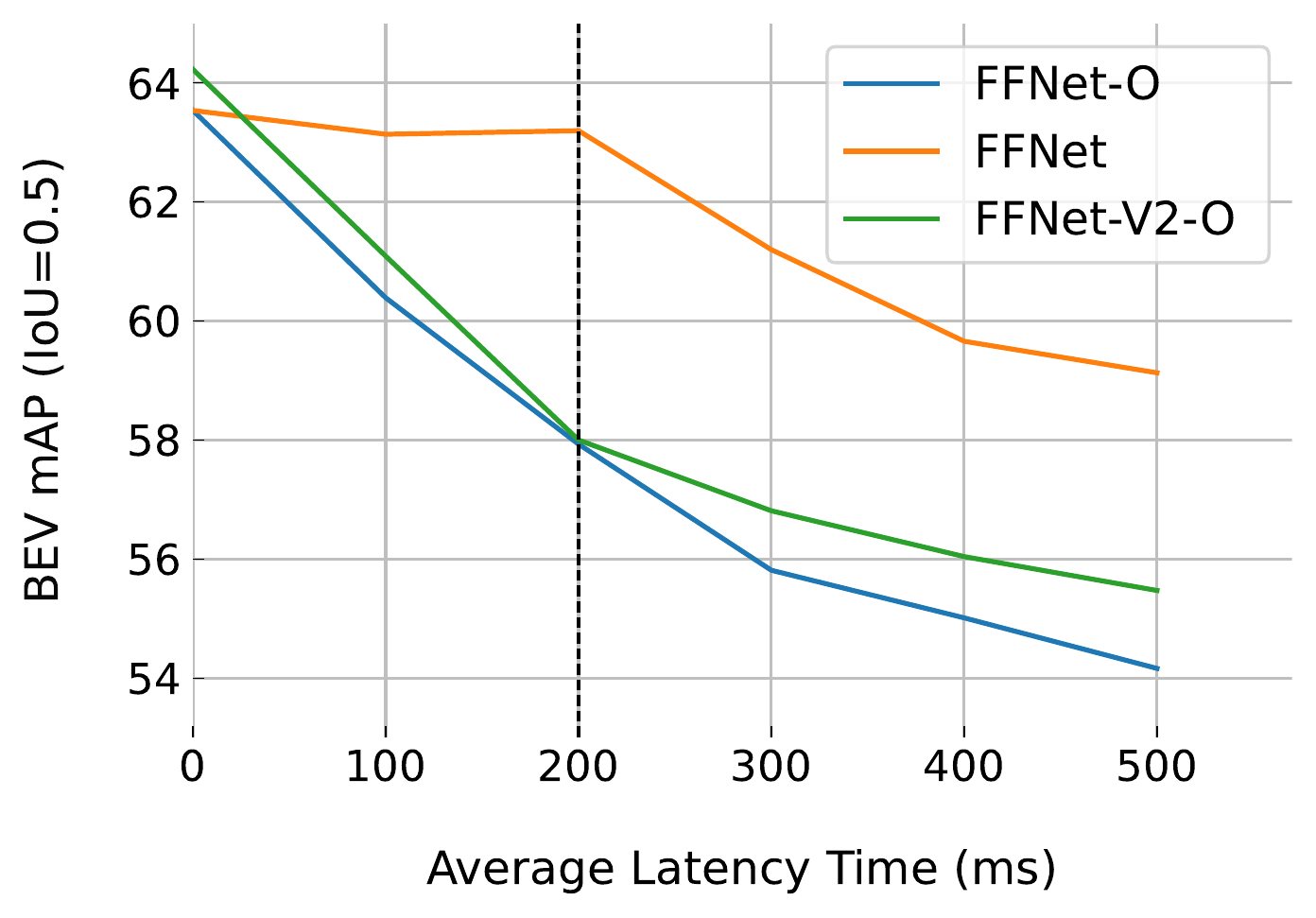}
\caption{\small Average latency vs. detection performance. FFNet-O denotes the FFNet (without prediction). FFNet-V2-O denotes the FFNet-V2 (without prediction). The yellow curve denotes the evaluation results of the FFNet, which behaves robustly across different latencies compared to models without prediction.
}\label{fig: different latency}
\end{figure}
We conducted further experiments to evaluate the performance of FFNet, FFNet (without prediction), and FFNet-V2 (without prediction) under various latency cases, ranging from 100$ms$ to 500$ms$. 
The experiment results are presented in Figure~\ref{fig: different latency}. 
As shown in the figure, both FFNet (without prediction) and FFNet-V2 (without prediction) exhibit continuous performance drops as the latency increases from 100$ms$ to 500$ms$. 
Specifically, in 500$ms$ latency, FFNet (without prediction) and FFNet-V2 (without prediction) exhibit a significant 9.38\% BEV-mAP (IoU=0.5) drop and 9.76\% BEV-mAP (IoU=0.5) drop, respectively. 
In contrast, FFNet shows little performance degradation within 200$ms$ latency and only has a 4.39\% BEV-mAP (IoU=0.5) drop.
These results indicate that FFNet is robust to different latency and can effectively handle the uncertain latency in VIC3D.

Robustness to uncertain latency is crucial because our feature flow could be received by different vehicles with varying latencies, as illustrated in Figure~\ref{fig: uncertain temporal asynchrony} in the Appendix. The feature flow must have the ability to make predictions at an arbitrary future time before being transmitted.
%%%%%%%%% conclusion
%%%%%%%%%%%%%%%%%%%%%%%%%%%%%%%%%%%%%%%%%%%%%%%%%%%%%%%%%%%%%%%%%%%%
%%%%%%%%%%%%%%%%%%%%%%%%%%%%%%%%%%%%%%%%%%%%%%%%%%%%%%%%%%%%%%%%%%%%
\section{Conclusion}
This paper proposes a novel intermediate-level cooperative framework called FFNet to address the object detection task in VIC3D. 
The proposed FFNet can effectively overcome the temporal asynchrony challenge and reduce the transmission cost by transmitting the compressed feature flow for cooperative detection. Additionally, a novel self-supervised learning approach is designed to train the feature flow generator.
Experimental results on the DAIR-V2X dataset demonstrate that the FFNet outperforms existing state-of-the-art methods. 
Notably, our approach emphasizes improving feature prediction quality and can complement existing intermediate-level methods. 
Furthermore, our approach can be extended to other modalities, including image and multi-modality data. 
The proposed method can also benefit multi-vehicle cooperative perception. 
Moreover, leveraging additional frames to generate feature flow can enhance the feature prediction ability of our method.
In summary, the proposed FFNet with feature prediction and self-supervised learning is a promising approach for VIC3D object detection, and can potentially be applied to various cooperative perception tasks in the future. 

\balance
\bibliography{reference}
\bibliographystyle{icml2023}

%%%%%%%%%%%%%%%%%%%%%%%%%%%%%%%%%%%%%%%%%%%%%%%%%%%%%%%%%%%%%%%%%%%%%%%%%%%%%%%
%%%%%%%%%%%%%%%%%%%%%%%%%%%%%%%%%%%%%%%%%%%%%%%%%%%%%%%%%%%%%%%%%%%%%%%%%%%%%%%
% APPENDIX
%%%%%%%%%%%%%%%%%%%%%%%%%%%%%%%%%%%%%%%%%%%%%%%%%%%%%%%%%%%%%%%%%%%%%%%%%%%%%%%
%%%%%%%%%%%%%%%%%%%%%%%%%%%%%%%%%%%%%%%%%%%%%%%%%%%%%%%%%%%%%%%%%%%%%%%%%%%%%%%
\newpage
\appendix
\onecolumn
\section{More Details about VIC3D Object Detection}
\subsection{Uncertain Temporal Asynchrony}
\begin{figure}[ht]
  \centering
  \includegraphics[width=1\linewidth]{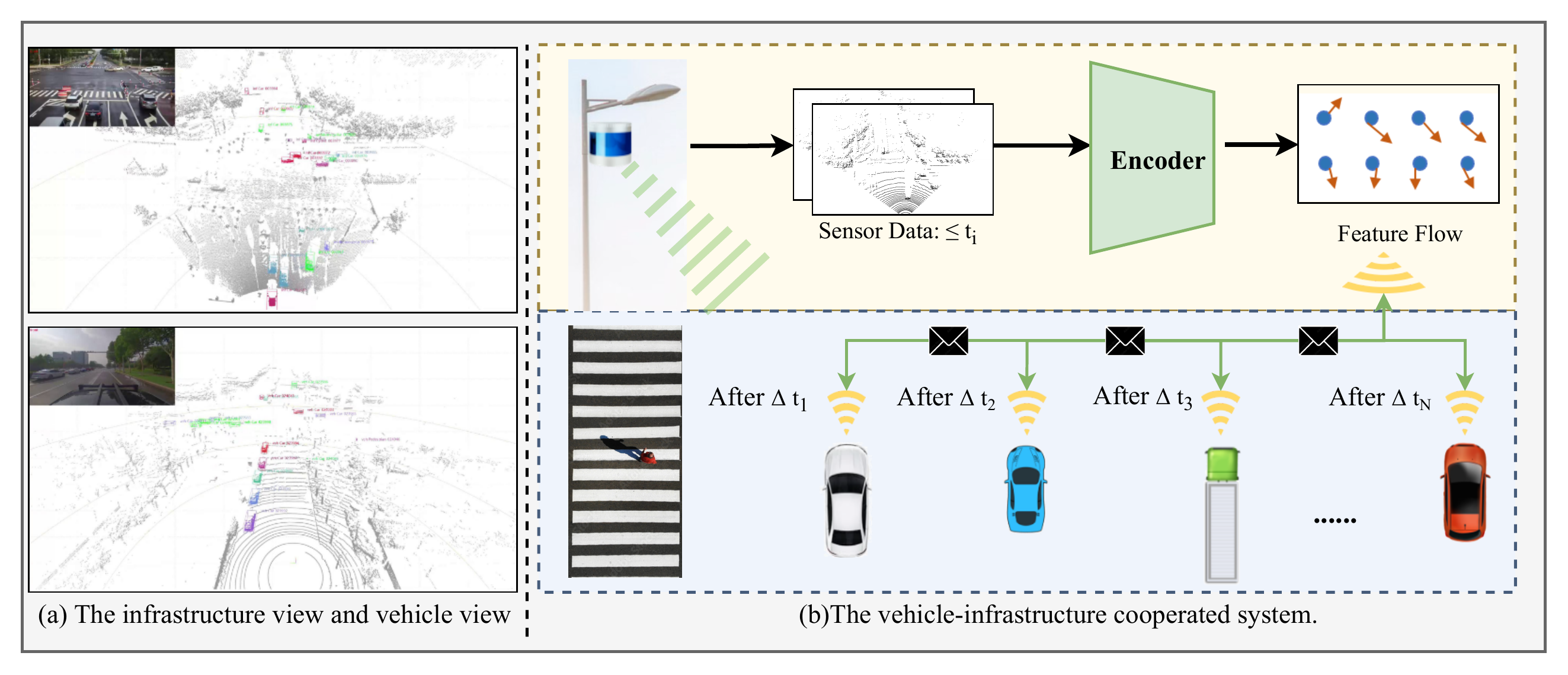}
  \vspace{-10pt}
  \caption{Uncertain Temporal Asynchrony In Vehicle-Infrastructure Cooperative 3D Object Detection.
  (a) The illustration depicts the utilization of infrastructure sensors to enhance the perception ability of autonomous driving vehicles by providing sensor data from different views. For instance, the infrastructure sensor data can be used to perceive objects located in the long-range zone of the vehicle sensor.
  (b) The vehicle-infrastructure cooperative 3D object detection system encounters uncertainty in communication delay $\Delta t$ occurring before the vehicle devices receive the infrastructure data. Due to limited communication conditions and varying ranges, the infrastructure data received by each autonomous driving vehicle exhibits varying latencies ranging from $\Delta t_1$ to $\Delta t_N$.
  }\label{fig: uncertain temporal asynchrony}
\end{figure}

In this section, we elaborate on the challenge of uncertain temporal asynchrony in the context of the VIC3D object detection task, which was briefly introduced in Section~\ref{sec: vic3d detection task}. In practical traffic environments, autonomous driving vehicles encounter limited communication conditions and varying ranges, resulting in the reception of infrastructure data with varying latencies. This introduces uncertainty in the temporal synchronization of the received data. We present a visualization example in Figure~\ref{fig: uncertain temporal asynchrony} to illustrate this challenge.
This uncertainty in temporal asynchrony is characterized by a significant range, which necessitates a flexible prediction strategy. To address this issue, we propose the extraction of feature flow from sequential infrastructure frames. This feature flow can be utilized to predict future features at any arbitrary future time, thereby providing the necessary flexibility to account for the uncertain temporal asynchrony. The predicted feature can then be used for cooperative detection, which can mitigate the fusion errors due to the uncertain temporal asynchrony.

\subsection{Detailed Evaluation Metrics for VIC3D}\label{sec: evaluation metrics}
This part provides a detailed explanation of the evaluation metrics used to assess the performance of the VIC3D object detection algorithm, namely mean Average Precision (mAP) and Average Byte ($\mathcal{AB}$).

\paragraph{Mean Average Precision (mAP).} We use mAP as a metric to evaluate the detection performance, as commonly used in previous works such as~\cite{yu2022dairv2x, everingham2015pascal}. The AP is calculated based on the 11-points interpolated precision-recall curve and is defined as follows:
\begin{equation}
\begin{aligned}
    AP & = \frac{1}{11} \sum_{r\in {0.0,...,1.0}} AP_{r} \
        & = \frac{1}{11} \sum_{r\in {0.0,...,1.0}} P_{interp}(r),
\end{aligned}
\end{equation}
where $P_{interp}(r) = \mathop{max}\limits_{\tilde{r}\geq r}p(\tilde{r})$,
and a prediction is considered positive if the Intersection over Union (IoU) is greater than or equal to 0.5 or 0.7, respectively. 
Here, we remove objects outside the egocentric surroundings, which we define as a rectangular area with coordinates [0, -39.12, 100, 39.12]. Finally, we average the AP across all classes to obtain the mAP.

\paragraph{Average Byte ($\mathcal{AB}$).}
We use $\mathcal{AB}$ as a metric to evaluate the transmission cost. 
In our implementation, we represent each element in the transmitted data using a 32-bit float, and we ignore the transmission cost of calibration files and timestamps.
We now explain how to calculate the transmission cost for each of the three transmission forms:
\begin{itemize}
    \item Calculating the transmission cost of transmitting raw data for early fusion: We represent each point in the point clouds with $(x,y,z, \text{intensity})$, and each point requires four 32-bit floats, equivalent to 16 Bytes. If there are 100,000 points in the point clouds, the $\mathcal{AB}$ of the transmission cost is $1.6\times 10^6$ Bytes.
    \item Calculating the transmission cost of transmitting detection outputs for late fusion: We represent each 3D detection output with $(x, y, z, w, l, h, \theta, \text{confidence})$, and each detection output requires eight 32-bit floats, equivalent to 32 Bytes. If we transmit ten detection outputs per transmission on average, the $\mathcal{AB}$ of the transmission cost is $3.2\times 10^2$ Bytes.
    \item Calculating the transmission cost of transmitting intermediate data for middle fusion: The feature or feature flow is represented with a tensor, and each element is represented by a 32-bit float. If the size of the feature or feature flow is $(100, 100, 100)$, then the $\mathcal{AB}$ of the transmission cost is $100\times 100\times 100\times 4$ Bytes, which is equivalent to $4\times 10^6$ Bytes.
\end{itemize}

\clearpage
\section{Implementation Details of FFNet}
\paragraph{Architecture.}
In this part, we provide a detailed architecture for FFNet, which is composed of the following six main parts.
1) The feature flow generation module. The infrastructure PFNet (Pillar Feature Net) shares the same architecture as PointPillars~\cite{lang2019pointpillars}.
The x, y, and z ranges of the input point cloud are [(0, 92.16), (-46.08, 46.08), (-3, 1)] meters, respectively.
The voxel sizes for x, y, and z are [0.16, 0.16, 4] meters, respectively.
The output shape of the pseudo-images is (64, 576, 576).
Both the infrastructure feature extractor $F_i(\cdot)$ and the estimated first-order derivative generator $\widetilde{F}_i^{'}(\cdot)$ use the same architecture of Backbone and FPN as SECOND~\cite{yan2018second}, with output shapes of [384, 288, 288].
2) The compressor and decompressor. 
The compressor uses four convolutional blocks with strides (2, 1, 2, 2) to compress the features from (384, 288, 288) to (384/32, 288/8, 288/8).
The decompressor uses three deconvolutional blocks with strides (2, 2, 2) to decompress the features back to the original size.
3) Affine transformation module. 
The affine transformation is implemented with the $affine_grid$ function supported in Pytorch.
We ignore the rotation around the x-y plane.
4) Feature fusion module.
The fusion module is a $3\times3$ convolutional block with stride 1 to compress the concatenated feature from (768, 288, 288) to (384, 288, 288).
5) Vehicle feature extractor. 
This extractor has the same configuration as the infrastructure PFNet and the feature extractor.
6) 3D object detection head.
We use a Single Shot Detector (SSD)~\cite{liu2016ssd} to generate the 3D outputs.
The anchor has a width, length, and height of (1.6, 3.9, 1.56) meters, with a z-center of -1.78 meters.
Positive and negative thresholds of matching are 0.6 and 0.45, respectively.

In this part, we provide a detailed architecture for FFNet, which is composed of following six main parts.
(1) The feature flow generation module: The infrastructure PFNet (Pillar Feature Net) shares the same architecture as PointPillars~\cite{lang2019pointpillars}. 
The x, y, and z ranges of the input point cloud are [(0, 92.16), (-46.08, 46.08), (-3, 1)] meters, respectively.
The voxel size of x, y, and z are [0.16, 0.16, 4] meters, respectively.
The output shape of the pseudo-images is (64, 576, 576).
The feature extractor $F_i(\cdot)$ and the estimated first-order derivative generator $\widetilde{F}_i^{'}(\cdot)$ both use the same Backbone and FPN as SECOND~\cite{yan2018second}, with output shapes of [384, 288, 288].
(2) The compressor and decompressor: The compressor uses four convolutional blocks with strides (2, 1, 2, 2) to compress the features from (384, 288, 288) to (384/32, 288/8, 288/8). The decompressor uses three deconvolutional blocks with strides (2, 2, 2) to decompress the features back to the original size.
(3) Affine transformation module: The affine transform is implemented with the $affine_grid$ function supported in Pytorch. Rotation around the x-y plane is ignored.
(4) Feature fusion module: The fusion module is a $3\times3$ convolutional block with stride 1 to compress the concatenated feature from (768, 288, 288) to (384, 288, 288).
(5) Vehicle feature extractor: This extractor has the same configuration as the infrastructure PFNet and feature extractor.
(6) 3D object detection head: A Single Shot Detector (SSD)~\cite{liu2016ssd} is used to generate the 3D outputs. The anchor has a width, length, and height of (1.6, 3.9, 1.56) meters, with a z-center of -1.78 meters. The positive and negative thresholds of matching are 0.6 and 0.45, respectively.

\paragraph{Training Settings.}
We trained the feature fusion base model on the training set of DAIR-V2X for 40 epochs with a learning rate of 0.001 and weight decay of 0.01. 
We randomly selected 11037 frames from the training set of DAIR-V2X and randomly set $k$ from the range [1, 2] to form $\mathcal{D}$. The explanation of $k$ and $\mathcal{D}$ can be found in Sec.~\ref{sec:semi-supervised-learning}. The trained feature fusion base model was used to pretrain FFNet.
We trained the feature flow generator on $D_u$ for 10 epochs, with a learning rate of 0.001 and weight decay of 0.01. All training and evaluation were implemented on an NVIDIA GeForce RTX 3090 GPU.

\clearpage
\section{Implementation Details of V2VNet and DiscoNet for VIC3D}
\paragraph{V2VNet for VIC3D.}
V2VNet~\cite{wang2020v2vnet} is a pioneering work in multi-vehicle cooperative perception, which is the first to transmitting intermediate-level data for cooperative perception without utilizing sequential frames for extracting temporal correlations. In this paper, we apply this approach to solving the VIC3D problem, as shown in Figure~\ref{fig: V2Vnet}. We made two modifications to the V2VNet architecture: (1) we removed multi-vehicle selection and kept only one vehicle settings as the infrastructure settings, and (2) we compressed the features from (384, 288, 288) to (384, 288/8, 288/8) to keep the transmission cost comparable to FFNet. The other modules keep the same configurations as their corresponding modules in FFNet.
We trained the V2VNet model on the training part of the DAIR-V2X dataset for 40 epochs, with a learning rate of 0.001 and weight decay of 0.01. The other training configurations are the same as those used for training FFNet.

\begin{figure}[ht!]
  \centering
  \includegraphics[width=1\linewidth]{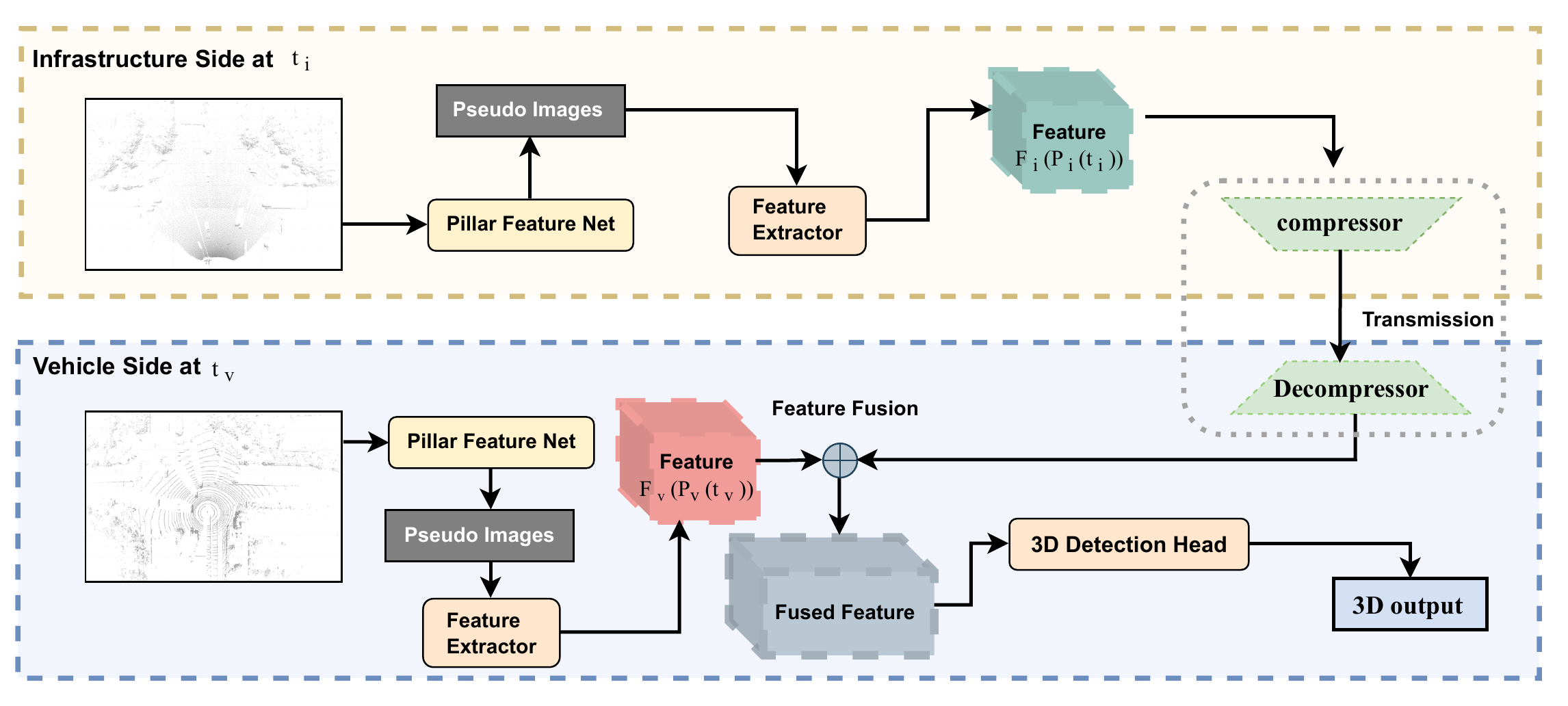}\label{fig: V2Vnet}
  \caption{\small \textbf{Implementation of V2VNet for Solving VIC3D Problem.} The V2VNet directly transmit the feature generated from the single point cloud, and the fuse it with the vehicle feature. This feature fusion could cause serious fusion error by the temporal asynchrony.}\label{fig: V2Vnet}
\end{figure}

\paragraph{DiscoNet for VIC3D.}
DiscoNet~\cite{li2021learning} was originally developed for cooperative perception among multiple vehicles. 
This method employs a teacher-student paradigm, where cooperative perception with raw data is used as the teacher network to instruct cooperative perception with intermediate data as the student network. To adapt DiscoNet to the VIC3D task, we use an early-fusion network as the teacher network and V2VNet as the student network. We train both the teacher and student models on the training subset of DAIR-V2X for 40 epochs, with a learning rate of 0.001 and weight decay of 0.01. Next, we fine-tune the student network using soft labels generated by the early-fusion network for an additional 10 epochs, with a learning rate of 0.0001 and weight decay of 0.01

\clearpage
\section{Comparison of Feature Flow Extraction on Different Sides}~\label{sec: extraction on different side}
This section discusses the effect of extracting the feature flow on different sides (infrastructure side \textit{vs.} vehicle side).

\begin{figure}[ht!]
  \centering
  \includegraphics[width=1\linewidth]{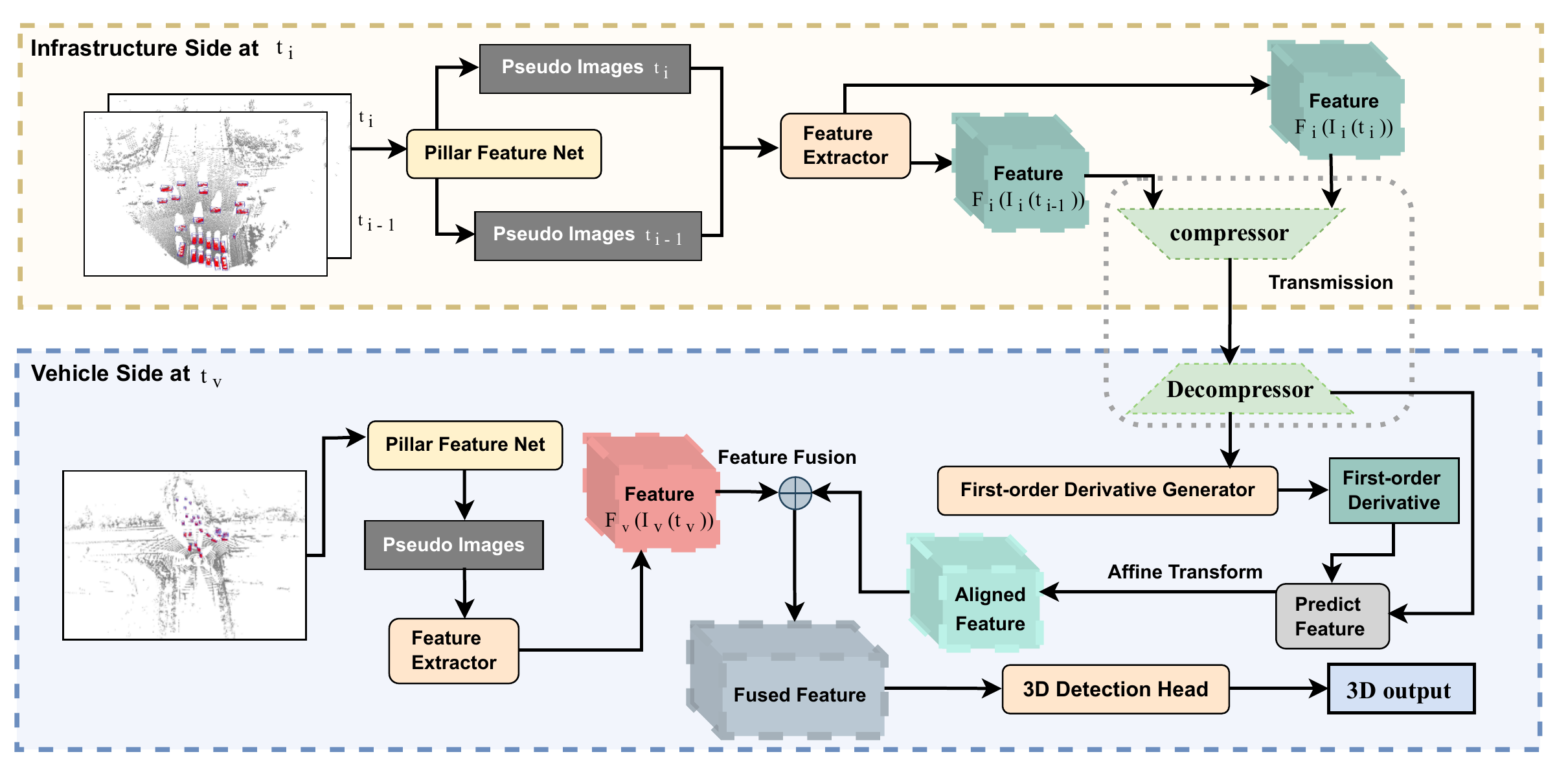}
  \vspace{-20pt}
  \caption{\small FFNet-V Overview. FFNet-V generates the feature flow on vehicle devices.}
\label{fig: FFNet-v}
\end{figure}

\paragraph{Experiment Setting.}
To compare the effect of feature flow extraction on infrastructure side and vehicle side, we train a modified FFNet called FFNet-V. The FFNet-V inputs the features $F_i(P_i(t_i-1))$ and $F_i(P_i(t_i-1))$ produced from consecutive infrastructure frames to generate the feature flow on vehicle devices.
We first concatenates the two received features and feeds them into a first-order derivative generator to generate the estimated first-order derivative of the feature flow $\widetilde{F}_i^{'}(t_i)$.
Then we predict the future feature as Equation~\ref{eq: predict F(t_i+k)}.
In addition, FFNet-V shares the same architecture modules and training configuration as FFNet.
The FFNet-V implementation framework is shown in Figure~\ref{fig: FFNet-v}.
To ensure a fair comparison with FFNet, we also train another FFNet-V by compressing the feature from (384, 288, 288) to (384/16, 288/8, 288/8), which has the same transmission cost as FFNet. This version of FFNet-V is referred to as FFNet-V (Same-TC). We evaluate both FFNet-V and FFNet-V (Same-TC) under different latencies (100$ms$, 300$ms$, and 500$ms$).

\paragraph{Result Analysis.}
Table~\ref{tab: flow extraction comparison} shows that FFNet-V, FFNet-V (Same-TC), and FFNet outperform FFNet-O under different latencies. This indicates that all these methods can reduce the detection performance drop caused by temporal asynchrony. 
However, FFNet can compensate for more performance drop and achieve better performance than FFNet-V and FFNet-V (Same-TC). 
Specifically, FFNet outperforms FFNet-V (Same-TC) by more than 3\% mAP@BEV (IoU=0.5) in 300$ms$ latency, while having the same transmission cost. The results demonstrate that extracting feature flow from raw sequential frames on infrastructure can improve the VIC3D detection performance more effectively than extracting feature flow from intermediate sequential feature frames on vehicle. 
Moreover, FFNet requires much fewer ego-vehicle computing resources, and the computing cost complexity (CCC) is only O(N), since the feature flow has already been produced on infrastructure devices and does not need to be generated on vehicle devices again. In contrast, FFNet-V consumes computation resources up to O(N) to extract flow from past features. Therefore, FFNet is more computation-friendly to resource-limited vehicle devices. Additionally, extracting feature flow on the vehicle requires much more storage because the feature flow extraction depends on the past frames that the vehicle received. Furthermore, FFNet-V relies heavily on past consecutive frames, so dropped frames can significantly affect the execution and performance. Therefore, FFNet is more storage-friendly to the ego vehicle and more robust to frame dropping.

\clearpage
\begin{table*}[h]
\small
\centering
\caption{\small
\textbf{Extracting Feature Flow on Infrastructure side \textit{vs.} on Vehicle Side.}
FFNet-O denotes the FFNet model without feature prediction.
FFNet-V denotes the model that extracts the feature flow on vehicle.
FFNet-V (Same-TC) denotes the FFNet-V which has the same transmission cost as FFNet.
``AB'' denotes the average byte used to measure the transmission cost. 
``SCC'' indicates the storage cost complexity for the vehicle devices to store the past frames, 
``CCC'' indicates the computing cost complexity for vehicle to extract the feature flow, 
``N'' indicates the number of historical structures to be used.
``/'' indicates the FFNet-O does not need infrastructure transmission and extra computation and storage.
The SCC of FFNet is O(1) because it does not need extra historical frames on vehicle devices. 
At the same time, the SCC of extracting feature flow on vehicle is O(N) because extracting feature flow on vehicle needs past frames received from infrastructure.
Moreover, FFNet achieves better detection performs, and this advantage becomes more pronounced~(+3\% mAP) when latency increases to 300$ms$.}\label{tab: flow extraction comparison}
\scalebox{0.97}{
\begin{tabular}{cc|lcllccc}
\hline
\multirow{2}{*}{Model} & \multirow{2}{*}{Latency (ms)} & \multicolumn{2}{c}{mAP@3D~$\uparrow$}                & \multicolumn{2}{c}{mAP@BEV~$\uparrow$} & \multirow{2}{*}{AB(Byte)~$\downarrow$} & \multirow{2}{*}{SCC~$\downarrow$} & \multirow{2}{*}{CCC~$\downarrow$}  \\ \cline{3-6} 
&    & IoU=0.5 & \multicolumn{1}{l}{IoU=0.7} & IoU=0.5    & IoU=0.7    \\
\hline \hline
FFNet-O            & 100              & 52.18   & 27.99                 & 60.39      & 49.14      & /  & /  & / \\
FFNet-V            & 100              & 53.21   & 28.43                 & 61.50      & 50.50      & 6.2$\times 10^4$  & O(N)  & O(N) \\
FFNet-V (Same-TC)  & 100              & 53.17   & 28.45                 & 62.44      & 51.68      & 1.2$\times 10^5$  & O(N) & O(N) \\
\textbf{FFNet (Ours)}        & 100              & 55.48   & 31.50                   & \textbf{63.14} (\textcolor{red}{+0.7})     & 54.28  & 1.2$\times 10^5$   & O(1) & O(1)
\\ \hline \hline   
FFNet-O            & 300              & 49.03   & 27.39                 & 55.81      & 47.28      & /  & /  & / \\
FFNet-V           & 300              & 50.81   & 28.45                  & 57.75      & 49.62      & 6.2$\times 10^4$  & O(N)  & O(N) \\
FFNet-V (Same-TC) & 300              & 50.5   & 28.25                  & 58.02      & 50.03      & 1.2$\times 10^5$  &  O(N) &  O(N) \\
\textbf{FFNet (Ours)}       & 300              & 53.46   & 30.42                   & \textbf{61.20} (\textcolor{red}{+3.18})    & 52.44  & 1.2$\times 10^5$     & O(1)  & O(1)  \\ \hline \hline
FFNet-O            & 500              & 47.49   & 27.01                 & 54.16      & 45.99      & /  & /  & / \\
FFNet-V            & 500              & 49.93   & 28.63                & 56.42      & 48.87      & 6.2$\times 10^4$ & O(N)  & O(N)  \\
FFNet-V (Same-TC)  & 500              & 49.98   & 27.7                 & 56.99     & 49.55      & 1.2$\times 10^5$ & O(N)  & O(N)  \\
\textbf{FFNet (Ours)}        & 500              & 52.08   & 30.11                   & \textbf{59.13} (\textcolor{red}{+2.14})      & 51.70  & 1.2$\times 10^5$  &  O(1)  &  O(1) \\ \hline \hline
\end{tabular}
}
\label{table_xiaorong}
\end{table*}

\clearpage
\section{Relationship to Other Existing Possible Solutions}\label{sec:relationship-discussion}
Compared with other possible existing solutions, FFNet provides a more applicable paradigm to implement the vehicle-infrastructure cooperative 3D object detection with the following advantages:
\begin{itemize}[itemsep=2pt,topsep=0pt,parsep=0pt]
    \item Better Performance-Bandwidth Balance: FFNet achieves a better performance-bandwidth balance compared with early fusion and late fusion methods. Unlike early fusion, FFNet transmits compressed intermediate data, thereby reducing transmission costs. Moreover, compared with late fusion methods, FFNet transmits more valuable information for the egocentric object detection.
    \item Overcoming Temporal Asynchrony Challenge: FFNet can address the temporal asynchrony challenge between the data captured from vehicle sensors and received from infrastructure sensors. Compared with middle fusion methods like V2VNet~\cite{wang2020v2vnet} and DiscoNet~\cite{li2021learning}, which only transmit features and do not consider temporal asynchrony problem, FFNet transmits the feature flow with the feature prediction ability. The feature flow can be used to generate future features that are aligned with the vehicle feature, which can help alleviate fusion errors caused by temporal asynchrony. Notably, the first-order derivative in the feature flow is an independent module, and it can also be applied to newer feature fusion methods to achieve lower transmission costs.
    \item Computing-friendly for Vehicles with Limited Computing Resources: Our feature flow is generated on the infrastructure side and can directly predict future features to compensate for uncertain latency with linear computation in ego vehicles. Another possible solution to solve the temporal asynchrony problem is proposed in~\cite{lei2022latency}, which generates future features with received historical features on vehicle devices. However, this solution requires significant computing resources to process the historical frames and extract temporal correlations to predict future features. Furthermore, extracting temporal information from compressed features could be challenging, as compressed features have lost much valuable information compared to raw sequential point clouds.
    \item Saving Annotation Costs: FFNet training saves much annotation costs. We use a self-supervised learning method to train the feature flow generator and extract temporal feature flow from sequential point clouds. Our training method does not rely on labeled data, and it is possible to exploit the massive unlabelled infrastructure-side sequential data in the future.
\end{itemize}

% \clearpage
% \section{Visualization Results}
% In this section, we provide more visualization results to illustrate the effectiveness of our FFNet.
% In this section, we present additional visualization results to further demonstrate the effectiveness of our proposed FFNet. 
% \begin{figure}[ht!]
%   \centering
%   \includegraphics[width=1\linewidth]{images/ffnet-effect.png}\label{fig: ffnet performance}
%   \caption{\small \textbf{Detection Performance of FFNet.} On the vehicle side, a car is occluded by other objects and cannot be seen from the vehicle view. On the infrastructure side, this occluded car can be seen from an infrastructure view. Our FFNet can detect this car.}
% \end{figure}

%%%%%%%%%%%%%%%%%%%%%%%%%%%%%%%%%%%%%%%%%%%%%%%%%%%%%%%%%%%%%%%%%%%%%%%%%%%%%%%
%%%%%%%%%%%%%%%%%%%%%%%%%%%%%%%%%%%%%%%%%%%%%%%%%%%%%%%%%%%%%%%%%%%%%%%%%%%%%%%

\end{document}